# Artificial Immune Systems

Julie Greensmith, Amanda Whitbrook and Uwe Aickelin

Abstract The human immune system has numerous properties that make it ripe for exploitation in the computational domain, such as robustness and fault toler- ance, and many different algorithms, collectively termed Artificial Immune Systems (AIS), have been inspired by it. Two generations of AIS are currently in use, with the first generation relying on simplified immune models and the second genera- tion utilising interdisciplinary collaboration to develop a deeper understanding of the immune system and hence produce more complex models. Both generations of algorithms have been successfully applied to a variety of problems, including anomaly detection, pattern recognition, optimisation and robotics. In this chapter an overview of AIS is presented, its evolution is discussed, and it is shown that the diversification of the field is linked to the diversity of the immune system itself, leading to a number of algorithms as opposed to one archetypal system. Two case studies are also presented to help provide insight into the mechanisms of AIS; these are the idiotypic network approach and the Dendritic Cell Algorithm.

## 1 Introduction

Nature has acted as inspiration for many aspects of computer science. A trivial ex- ample of this is the use of trees as a metaphor, consisting of branched structures, with leaves, nodes and roots. Of course, a tree structure is not a simulation of a tree, but it abstracts the principal concepts to assist in the creation of useful computing systems. Bio-inspired algorithms and techniques are developed not as a means of simulation, but because they have been inspired by the key properties of the un- derlying metaphor. The algorithms attempt to improve computational techniques by mimicking (to some extent) successful natural phenomena, with the goal of achiev-

Julie Greensmith, Amanda Whitbrook and Uwe Aickelin
School of Computer Science, University of Nottingham, Nottingham, NG8 1BB, UK e-mail: {jqg, amw, uxa}@cs.nott.ac.uk

ing similar desirable properties as the natural system. This is demonstrated in both neural networks [17] and genetic algorithms [25].

Artificial Immune Systems (AIS) [20] are algorithms and systems that use the human immune system as inspiration. The human immune system is a robust, decentralised, error tolerant and adaptive system. Such properties are highly desirable for the development of novel computer systems. Unlike some other bio-inspired techniques, such as genetic algorithms and neural networks, the field of AIS encompasses a spectrum of algorithms that exist because different algorithms implement different properties of different cells. All AIS algorithms mimic the behaviour and properties of immunological cells, specifically B-cells, T-cells and dendritic cells (DCs), but the resultant algorithms exhibit differing levels of complexity and can perform a range of tasks.

The major part of AIS work to date has been the development of three algorithms derived from more simplified models; negative selection, clonal selection and im- mune networks. However, these first-generation AIS algorithms have often shown considerable limitations when applied to realistic applications. For this reason, a second generation of AIS is emerging, using models derived from cutting-edge immunology as their basis, not simply mechanisms derived from basic models found in text books.

The aim of this chapter is to give the reader an overview of the field of AIS by taking a high level perspective of its evolution. In section 2 an overview of the major developments in immunology is presented, incorporating a number of immunologi- cal theories. Section 3 describes the development of AIS over the past two decades, and the next two sections showcase two particular examples of AIS algorithms; an idiotypic network in section 4 and the Dendritic Cell Algorithm (DCA) in section 5. Section 6 concludes the chapter, with a summary and details of potential future trends.

## 2 Immunological Inspiration

The human immune system can be used as inspiration when developing algorithms to solve difficult computational problems. This is because it is a robust, decen- tralised, complex, and error-tolerant biological system; i.e. it possesses properties that make it ideal for certain application areas, such as computer intrusion detection and pattern recognition. The human system is also well-studied within immunology, and is viewed as the most sophisticated of immune systems in nature. Although its precise function remains undetermined, it is postulated that it has two roles; to pro- tect the body against invading micro-organisms (pathogens), and to regulate bodily functions (homeostasis).

Immunologists like to describe the immune system as consisting of two parts, namely the innate immune system and the adaptive immune system. It was origi- nally thought that these were two distinct sub-systems with little crossover, with the innate system responding to known threats and the adaptive immune system tack-ling previously un-encountered threats. However, current research suggests that it is the interplay between these two systems that provides the high level of protection required, i.e. the ability to discriminate between 'self' and 'nonself' entities.

In this section the basic principles of immunology are introduced from the historical perspective of their development. For a more comprehensive, biological view of the immune system, the interested reader should refer to any of a number of more standard immunology texts, for example [43] and [15].

## 2.1 Classical Immunology

Until relatively recently, the central dogma of immunology was self-nonself discrimination through the principles of clonal expansion and negative selection. These concepts have dominated the field since they were first described, as they provide adequate explanation of the function of the adaptive immune system over the life-time of an individual.

In 1891, Paul Ehrlich [49] and his colleagues postulated that the defence mechanism against pathogens was the generation of immunity through the production of immunoglobulins termed antibodies. They showed that the antibodies generated are specific to the pathogen (antigen) being targeted, and suggested that the immune system must remove these antigens before an infection spreads, without responding to its own cells. This led to the theory of the horror autotoxicus, which states that "an organism would not normally mobilise its immunological resources to effect a destructive reaction against its own tissues" [48]. It was later discovered that a par-ticular type of lymphocyte (white blood cell) termed a B-cell is responsible for the production of antibodies, and that the antibodies are proteins that can potentially bind to the proteins present on the invading antigens.

Following the characterisation of antibodies, the theory of clonal selection was proposed by Burnet [48]. This mechanism corroborated the notion of horror autotoxicus and found that "an individual somehow manages to prevent all future ability to respond to auto antigens i.e. self, leaving intact the ability to respond actively to the universe of other antigens i.e. nonself" [48]. The notion implied that im-mune function contains a mechanism of tolerance, which Burnet described as an "irreversibly determined immunological self". This formed a major constituent of a theory known as central tolerance and was subsequently proven as valid experi-mentally, earning Burnet a Nobel prize for his efforts.

The clonal selection theory has two constituents. First, B-cells are selected to be fit for purpose during a 'training period'. Cells expressing receptors (cell surface-bound antibodies) that can match antigen are kept to form the B-cell population, but cells that cannot bind to antigens are removed. Once B-cells are released into the periphery, encounter with external antigens causes the cells to produce free ver-sions of the B-cell receptor, i.e. antibodies, which can bind to the matching antigen. Second, the process of antibody tuning occurs through somatic hypermutation and affinity maturation [46]. If a B-cell matches an external antigen, the cell clones it-self. However, the hypermutation process ensures that exact clones are not formed; the clones express B-cell receptors that are slight variants on the parent cell's re- ceptor. This is a type of biological optimisation, ultimately resulting in antibodies that can bind more successfully to external antigens. The antibodies can therefore be used as markers of nonself entities within the body. The whole process is termed affinity maturation and is used to generate the most responsive antibodies.

As the century progressed, a second class of white blood cell, T-cells, were char- acterised, and in 1959, the principle of negative selection was proposed by Joshua Lederberg, a then colleague of Burnet. He established the link between foetal devel- opment and the generation of tolerance to self-substances, termed self-antigen, not- ing the co-occurence of the initial production of T-cells and tolerance to self-antigen. This led to the idea that the selection process implied "self learning through negative selection", and caused Lederberg to suggest that "whenever produced, lymphocytes (T-cells) undergo a period of immaturity during which antigen recognition results in their death" [16]. He also proposed that further activation of the T-cells in the tissue is needed for the cells to develop the ability to remove pathogens such as bacterial agents and virally-infected self cells.

During embryonic development in the womb, T-cells migrate to an immune sys- tem organ, the thymus. Whilst in the thymus, the newly created T-cells are exposed to a comprehensive sample of self-antigen. Any T-cell displaying a receptor which matches a self-antigen is removed. This process continues until puberty, after which the thymus shrinks to a negligible size. So-called 'self-reactive' T-cells are thus eliminated through this filtering process.

## 2.2 The Immunologists' 'Dirty Little Secret'

According to Ehrlich's horror autotoxicus, the immune system should not respond to self and should aim to eliminate all sources of nonself. However, this phenomenon is not always observed, and numerous noteworthy exceptions have been discov- ered [43], questioning the credibility of the self-nonself dogma, for example:

1. Vaccinations and immunisations require adjuvants, namely microbial particles that provide additional stimulation of the immune system;
2. What the body classes as self changes over time, an effect termed changing self. This phenomenon is observed in women during pregnancy;
3. Human intestines are host to colonies of bacteria that serve a symbiotic function. These organsims are clearly nonself, yet no immune response is mounted;
4. In the western world, an individual's immune system can sometimes start to re- spond to benign particles such as pollen, cat saliva, latex, peanut proteins etc. resulting in allergic reactions;
5. An individual's immune system can sometimes begin to attack its host in the form of autoimmune diseases, for example multiple sclerosis and rheumatoid arthritis.

## 2.3 Costimulation, Infectious-nonself and The Danger Theory

Three main theories have both challenged and augmented the process of self-nonself discrimination including:

- Costimulation
- Infectious nonself
- Danger signal recognition

Some of the cells involved in these theories are part of the innate immune system that was first observed by Metchikoff in 1882 [48]. He noted that invertebrates such as shrimp and starfish mobilise phagocytes, which ingest invading pathogens, clear-ing the threat from the host. This first line of defence is also found in humans, and consists of a diverse array of interacting cell types. The innate system was initially seen as the adaptive system's lesser counterpart, as it did not appear to be as sophis-ticated. However, there has been renewed interest in the innate system, as it is now thought to provide some of the answers to the problems associated with the theories of adaptive immunity.

The concept of costimulation was introduced in an attempt to overcome a prob-lem observed in the hypermutation of antibodies, i.e. if the resulting hypermutated antibodies have a structure that could react to self cells, it would cause horror auto-toxicus. It was hence suggested that B-cells work in conjunction with T-cells [48], and that a B-cell would be eliminated if it did not receive a costimulatory signal from a 'helper T-cell'. Later it was shown that helper T-cells are also regulated by a 'stimulator cell' that provides the costimulatory signals. These professional antigen-presenting cells are known as dendritic cells (DCs) and are part of the innate immune system. The process of costimulation casts doubt on the theory of central tolerance, placing the innate system in control of the immune response.

The infectious nonself model proposed by Janeway in 1989 [34] further improved understanding of costimulation. Janeway suggested that the DCs perform their own version of self-nonself discrimination. This is based on their ability to recognise the signatures of bacterial presence innately, a skill developed over millennia throughout the evolution of the species. It is shown that DCs contain a repertoire of receptors on their surface, tuned for binding to molecules produced exclusively by bacteria. These molecules are collectively termed PAMPs (pathogen-associated molecular patterns). Janeway showed that the induction of an immune response is facilitated by the production of costimulatory molecules from DCs. When exposed to PAMPs and antigen, the DC produces a collection of molecules that assist in their binding to a T-cell, increasing the time a T-cell remains in contact with a presented antigen. This timing issue is thought to be crucial in the activation of T-cells.

Infectious nonself can explain the need to add adjuvants to vaccines. Adjuvants are formed from neutered bacterial detritus, which, according to the theory, provide the PAMPs necessary to mount an immune response. It also explains why no re-sponse is mounted to changing self, as the absence of a second signal leads to the deactivation of T-cells. However, the infectious nonself model cannot explain toler-ance to symbiotic bacteria, which produce PAMPs, yet are not eradicated from the body. Furthermore, this model cannot explain the phenomena of autoimmunity and its relatively high frequency of occurrence in the western world.

Despite the addition of a second costimulatory signal to the self-nonself model, it became apparent that a piece of the immunological puzzle was still missing. It was unclear why the immune system should respond to self, or why bacteria producing PAMPs were not classed as foreign. In 1994, Matzinger proposed that the immune system is controlled by the detection of damage to the body [42], not the detection of specific antigen structures or bacterial products. Matzinger suggested that the activating danger signals do not come from external sources, but are produced by the cells of the body when a cell dies unexpectedly (necrosis). The danger theory also proposes that the cells of the innate immune system can actively suppress an immune response in the absence of danger and in the presence of molecular signals produced when cells die normally (apoptosis).

DCs are sensitive to both the signals of necrosis and apoptosis in addition to PAMPs, and are attracted to areas in which cells are dying. They collect debris, including potential antigens, and all of the molecules found in the extracellular matrix (their environment) contribute to the regulation of their internal signal processing mechanism. If a DC is exposed to the molecules from necrosing cells, it transforms to a mature state. If it is exposed to the suppressive molecules of apoptosing cells, then it is transformed to a semi-mature state. The DC eventually complexes with a T-cell, i.e., a DC and T-cell bind if the antigen collected and presented by the DC has a sufficient binding affinity with the T-cell antigen receptor. If the DC is in the mature state, the T-cell becomes activated and all entities bearing that antigen are eliminated. If the DC is semi-mature, the T-cell is tolerised to the presented antigen and no response to it is generated. In this way, the processing of the input molecular signals provides the immune system with a sense of context; in other words, if an entity is foreign but harmless, then the immune system does not waste resources responding to it.

The peripheral-tolerance danger model can also account for the effects of autoimmunity; when a self-protein is present in the same place and at the same time as the antigen of a pathogen, the immune system may respond to its own tissue, as both host and foreign antigens are collected by the same DCs. This has been framed within the context of multiple sclerosis, as the symptoms frequently appear in combination with bacterial or viral infection.

Despite its ability to explain several key anomalies, acceptance of the danger theory has been slow within immunology. There has been a lack of experimental evidence to support Matzinger's ideas, and no single 'danger signal' has been dis- covered, though characterisation of the molecules involved is improving as molec- ular techniques advance.

## 2.4 Idiotypic Networks: Interantibody Interactions

In addition to the research on mechanisms of immune discrimination, theories ex- ist that attempt to explain the various emergent properties of the immune system. One of these theories is the idiotypic immune network theory, initially proposed by Jerne in 1974 [35]. The theory postulates that interactions between immune cells (and not necessarily external agents) cause modulation in the behaviour of the im- mune system as a whole. This modulation is proposed to lead to the generation of immune memory, i.e. the ability of the immune system to remember past encoun- ters with pathogens, and hence provide a secondary response that is both accurate and rapid. The idiotypic network model does not attempt to contradict the principles outlined in classical immunology, but provides a complementary theory of antibody stimulation, where antibodies can influence other antibodies in addition to antigens. Idiotypic models have been developed, although no physical evidence exists to sup- port the theory.

## 2.5 Summary

Immunologists classify the human immune system into two distinct sub-systems, the innate and adaptive. Until recently the adaptive system, responsible for modifi- cation of the immune response over the lifetime of an individual (through the tuning of B and T-cells), was viewed as far more sophisticated than its innate counterpart. The selection mechanisms of the B and T-cells, and their processes of adaptation form the major part of the self-nonself principle, which states that the immune sys- tem is activated in response to the detection of foreign antigen, but does not respond to self antigen. However, the adaptive model of immune activation has problems associated with it, and these have led immunologists to look in greater detail at the innate immune system, adapted over the lifetime of the species, which responds quickly to invaders based on receptors encoded within the genome. It is now thought that it is the interaction between the innate and adaptive systems and their cells that provides the necessary protection, and consequently, there has been fresh interest in the cells of the innate system, for example DCs. These are responsible for translating and integrating information from the tissue to the T-cells, which results in either ac- tivation or tolerisation of the immune system. While the classical self-nonself view is important to immune function, the interplay between the two systems and the cor- responding cells influences the ultimate decision as to whether or not to respond to an antigen.

## 3 The Evolution of Artificial Immune Systems

AIS is the collective name for a number of algorithms inspired by the human im- mune system. Unlike genetic algorithms, for which there is an archetypal algorithm and variants thereupon, there exists no single algorithm from which all immune al- gorithms are derived. However, all research within AIS stems from foundations in theoretical immunology, and numerous parallel streams of research have been con- ducted over the past 20 years, resulting in the development of distinct sub-streams, including computational immunology. The evolution of the various approaches that exist within AIS is depicted in Figure 1, which shows significant papers (given in quotes) or algorithms that have shaped the field of AIS. The white-ringed hubs rep- resent the significant works within a particular sub-stream, and the terminating rect- angles show branches of the research that are not currently active. In addition, the proximity of the sub-stream to the stream of theoretical immunology in the cen- tre represents the extent of the immunological modelling, with more modern AIS approaches closer to the underlying metaphor.

The diagram also shows that AIS are classified into two distinct groups; first and second-generation algorithms. The first-generation algorithms use simplistic models of immunology as the initial inspiration, for example negative and clonal selection. In contrast, the second-generation algorithms, for example, the Dendritic Cell Algo- rithm (DCA) [29], are built on a foundation of interdisciplinary research that allows for a much finer-grained encapsulation of the underlying immunology. Although most of the second-generation algorithms are still in their infancy, and require much more theortectical study, they are showing great promise in a number of application areas.

In this section each of the sub-streams and its applications are described indi- vidually and in chronological order, so that the evolution of AIS can be traced. In particular, negative selection, clonal selection and immune network approaches (the key first-generation algorithms) are discussed in detail, and the recently-developed second-generation algorithms that use the 'Conceptual Framework' methodology, are also treated.

### 3.1 Computational and Theoretical Immunology

A vein of computational and theoretical immunology lies at the core of AIS, as the process of developing mathematical models of immunological mechanisms is similar, at least, in principle to the development of immune inspired algorithms. It is not surprising, therefore, that theoretical models of immune phenomena acted as a foundation for the initial AIS algorithms, clonal and negative selection, and immune network-based approaches.

In the case of the clonal selection principle, this was initially based on works carried out in the 1970s by Burnett [12], where affinity metrics were first charac- terised mathematically. In combination with this model, Jerne's idiotypic network model was formalised by Farmer et al. in the 1980s [22], and stipulated the interaction between antibodies mathematically. The network model was seen as having computationally useful properties, and provided a network-based approach distinct from both neural networks and genetic algorithms. The model was also interpreted by Bersini and Varela [9] with numerous refinements, and the combination of these two approaches forms the cornerstone of all AIS work that abstracts the idiotypic network.

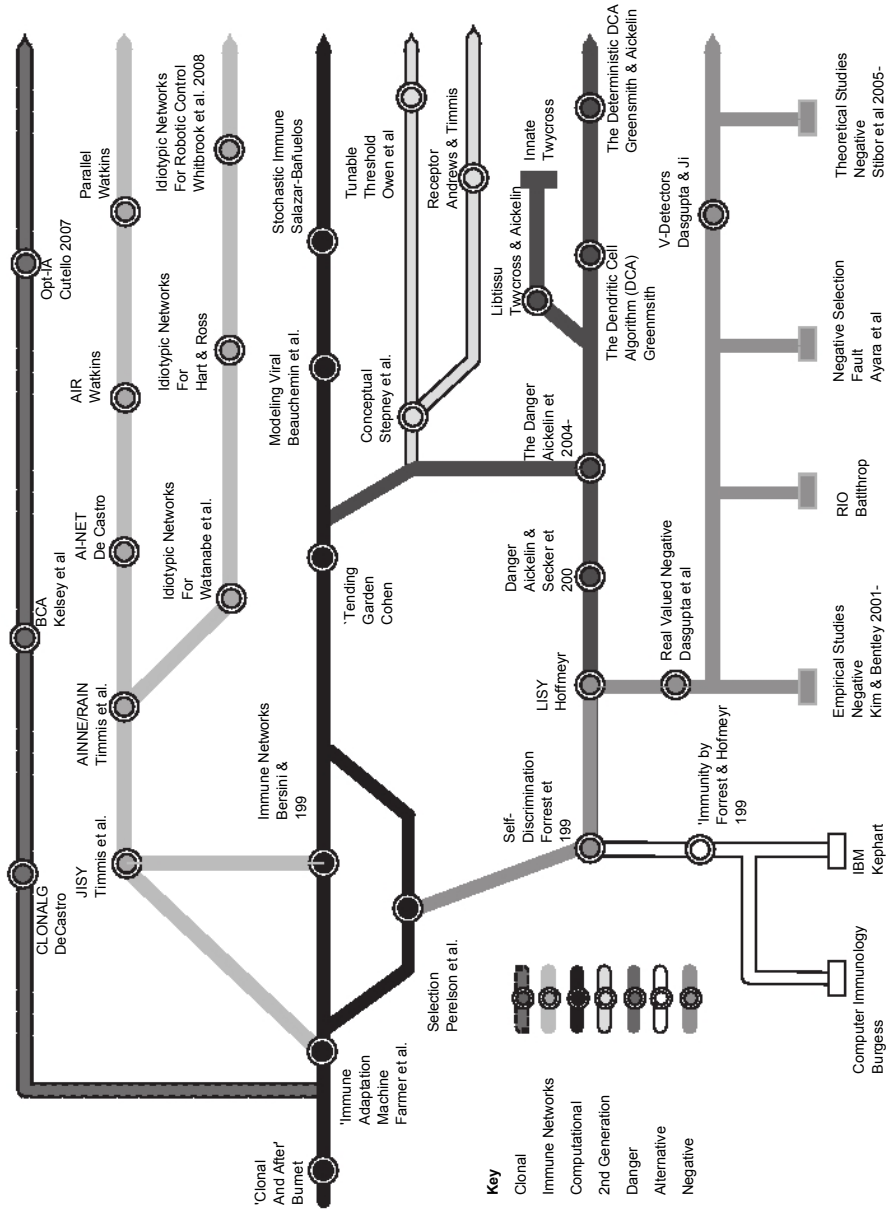

Fig. 1 The evolution of AIS from 1978 to 2008.

Similarly, a theoretical model of the selection of T-cells in the thymus by Perelson et al. [10] resulted in the development of negative selection as a technique within AIS. This model detailed the selection of T-cells (based on affinity metrics) to model the suitability of a T-cell receptor (TCR) for the detection of potential non-self anti- gen. The transfer from theoretical immunology to an AIS algorithm was by virtue of a collaboration between Perelson and Forrest, using Forrest's expertise in machine learning to improve Perelson's model.

It was around this time that the primary algorithms were applied to a battery of computational tasks, and AIS began to diverge from theoretical immunology. How- ever, although the initial performance of the developed algorithms was good, the techniques proved no better than the state-of-the-art algorithms that already existed in the chosen problem domains. Consequently, AIS researchers started to turn back to the underlying immunology (both experimental and theoretical), as it was as- sumed that the developed algorithms were based on out-dated, oversimplified mod- els of the computation actually performed by the human immune system. Of course, theoretical immunology had also progressed since the 1990s.

In 2004, Cohen published a book entitled 'Tending Adam's Garden' [14], that described the immune system as a complex adaptive system. Other similar research into a systemic perspective of the immune system, paired with the increase in pop- ularity of interdisciplinary approaches, enticed AIS researchers to renew their in- terest in the underlying metaphors. At this point, theoretical immunologists were welcomed into the field of AIS, acting as translators between the complicated and dynamic world of experimental immunology and computation. The AIS algorithms developed as a result of this incorporated many new ideas from modern immunol- ogy and show promise to out-perform older systems. In addition, in a bid to attract more researchers with a background in immunology, the AIS community devised a computational immunology stream as part of its conference [8]. Three examples of high quality research in this area include a model of viral dynamics [7], an in- vestigation into the cellular maximal frustration principle [1], and a model of the stochastic nature of immune responses [47].

As computational and theoretical immunology become more sophisticated, it seems likely that the boundaries between the two fields will blur, resulting in the development of more sophisticated AIS algorithms. AIS practitioners are hopeful that any new systems developed will remain faithful to the underlying principles, as stipulated by the creators of the 'Conceptual Framework' [51] approach to AIS development. Whether this approach will bear fruit is conjecture, but it has certainly given a lease of life to a field that has strayed far from its initial roots.

## 3.2 Negative Selection Approaches

The first example of an implemented AIS performing a useful computational task was an incarnation of a self-nonself discrimination system, used for the detection of computer virus executables [24]. (Incidentally, the precursor to this system was the original collaboration between computer science and immunology, i.e. the develop- ment of a genetic algorithm-based approach for understanding the mechanisms of pattern recognition within the immune system [50].) The self-nonself discrimina- tion system involved creating a behaviour profile of sequences of system calls on a computer network during a period of normal function. To aid in detecting malicious intruders, any subsequent sequences were matched against the normal profile, and any deviations reported as a possible intrusion. This research and its related work is perhaps the most widely known and popularised AIS to date [23], as the data used
is popular amongst the intrusion detection community, with nearly 1000 citations.

The approach attracted a great deal of attention from the security community, as exemplified in the research of Kephart [38]. This was the first attempt to apply AIS within a commercial setting, and consequently introduced AIS to a wider audience. The research was inspired by the efforts of Forrest et al. [23] and subsequent work by Hofmeyr and Forrest [32] in their paper 'Immunity by Design', and inspired a more systemic approach to AIS development, as pursued by Burgess [11]. Kephart also attempted to build on the system-profiling approach to intrusion detection by implementing a heterogeneous AIS. His work, and also that of Burgess, are good examples of alternative approaches based on the self-nonself principle.

However, the major development in this sub-stream was the introduction of a true negative selection algorithm in a system named 'Lisys', which consisted of three phases. Here, the first phase was used for the definition of self, i.e. the normal profile was generated from input data to encompass normal behaviour patterns de- fined in advance to form a sense of self. The second phase involved the generation of a set of random detectors containing a representation suitable for matching the patterns used to create the self profile. The final phase implemented the detection of anomalies in previously-unseen data by comparing each detector against all self patterns contained within the self profile. If any of the randomly-generated detectors matched a self pattern, the detector was deemed unsuitable and was removed from the detector set. However, if the detector did not match any self items it was saved and became part of the pool used for anomaly detection. Thus, when the highly- tuned detector set was presented with unseen test data, if any detector matched a pattern, the pattern was classed as anomalous and marked accordingly. A depiction of the algorithm at the core of this system is given in Figure 2.

A full description of the multiple-stage negative selection algorithm is given in the work by Hofmeyr [31], which encoded the detectors as bit strings and used an r-contiguous bit function for matching. Extensions to the work include the incorporation of real values into the encoding, the use of multi-dimentional vector representations known as V-detectors, and the use of adaptable thresholds to reduce false positives [6]. Negative-selection algorithms have also been employed to solve fault tolerance problems, and numerous other anomaly detection problems.

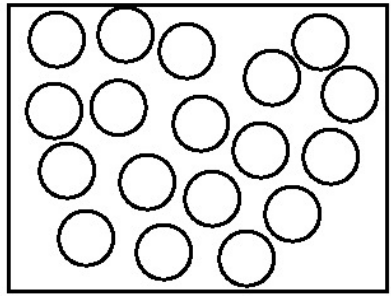

**Step One:**
Randomly generate initial detector-population with *n* detectors to cover the feature space, where each is  one detector.

**Step Two:**
Using the training data, define regions of `self' space, representing normal.

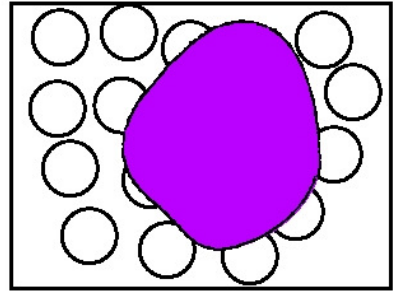

**Step Three:**
Delete all detectors which overlap with the defined self region, leaving detectors primed to detect nonself entities.

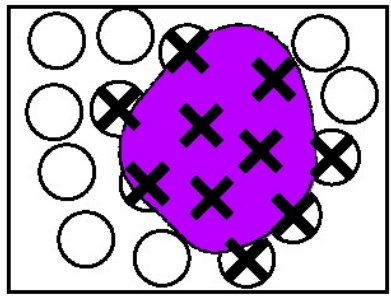

**Step Four:**
Introduce new pattern (antigen) and calculate affinity with nearest detector. If affinity is greater than a defined threshold, the detector is activated and the antigen is classed as anomalous ( 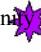 ). Antigen with insufficient affinity are classed as normal ( 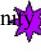 ).

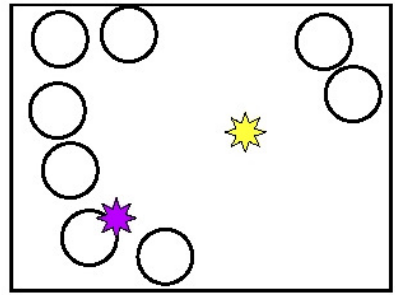

Fig. 2 An illustration of negative selection.

Despite its initial promise, negative selection has been shown to have a number of associated problems that can render it somewhat undesirable for use in network intrusion detection. First, the necessity to create a randomly-generated initial de- tector population can be prohibitive, because, as the dimensionality of the feature space increases, the number of detectors required to cover it increases exponentially. Second, negative selection is a one-shot supervised learning algorithm, where the definition of normal is not updated as time progresses. This is particularly relevant to computer security where what is defined as normal has the tendency to change over time. Negative selection algorithms can therefore cause excessive numbers of false positive alerts, which can cripple a system. The problems with the algorithm are discussed further in Kim and Bentley [39] and are proven theoretically by Stibor et al. [52]. Although numerous modifications and variants in representation have been made, such as the addition of variable length detectors, the algorithm seems fit for purpose only for small, constrained problems where the definition of normal is not likely to change and the set encompassing normal is small. For a comprehensive overview of the negative selection algorithm, the interested reader should refer to the review by Ji and Dasgupta [36].

### 3.3 Clonal Selection Approaches

During the early years of AIS, researchers recognised that, in addition to the T-cell inspiration employed by Forrest et al. [32], basic models of B-cells and their corre- sponding antibodies could act as a good underlying metaphor. B-cells produce an- tibodies of a specific configuration, and their diversity is stimulated upon encounter with a foreign antigen, where the resulting B-cell clones vary the receptor configura- tion in order to perform a biological local search to find the best-fitting receptor. The B-cell model appeared ripe for exploitation, given the similarities with local search and optimisation techniques, and in 2000 a theoretical model of the hypermutation process proposed by Burnett [12] served as inspiration for CLONALG [21], a pop- ular AIS algorithm involving an abstract version of the cloning and hypermutation process.

All clonal selection-based algorithms (CSA) essentially centre around a repeated cycle of match, clone, mutate and replace, and numerous parameters can be tuned, including the cloning rate, the initial number of antibodies, and the mutation rate for the clones. CLONALG, AI-NET, the B-cell algorithm [37] and AIRS [60] all incorporate this basic functionality. (AI-NET contains constituents of both CSA and immune network approaches [19].) The CSA used in CLONALG is illustrated in Figure 3.

CSAs have a strong resemblance to genetic algorithms without crossover, but their notion of affinity and their significantly higher mutation rate (the hypermu- tation component) distinguish them from similar adaptive algorithms. CSAs also share properties with both K-means and K-nearest neighbour approaches. The CSA technique would be most similar to a K-nearest neighbour scheme where K is one,

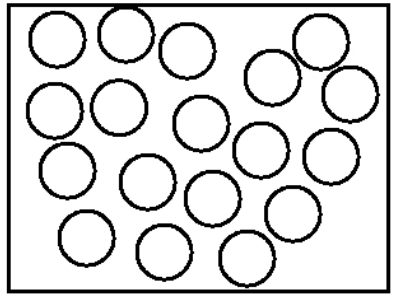

**Step One:**
Randomly generate initial antibody population with *n* detectors, where each  is one antibody.

**Step Two:**
Introduce new pattern (antigen ) and select the nearest clone (coloured) using a defined distance metric such as the Euclidean distance.

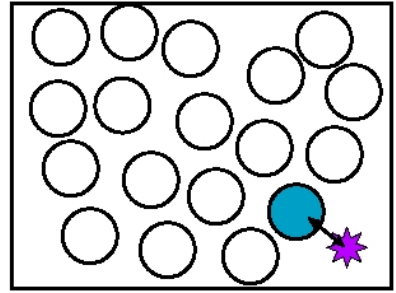

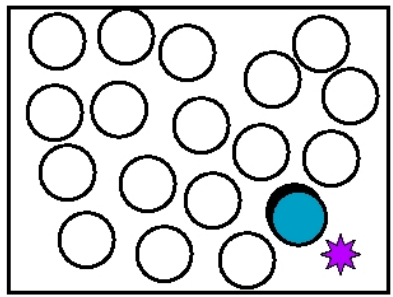

**Step Three:**
Clone nearest antibody in proportion to the affintiy between antibody and antigen. The greater the affinity the greater the number of clones produced.

**Step Four:**
Mutate clones, with distance of mutation inversely proportional to affinity. The greater the affinity the greater the distance between mutant antibodies.

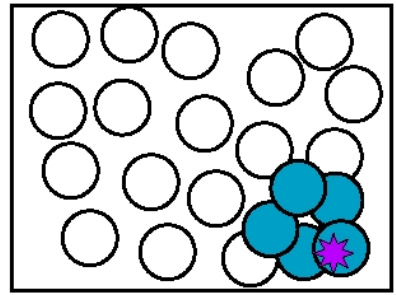

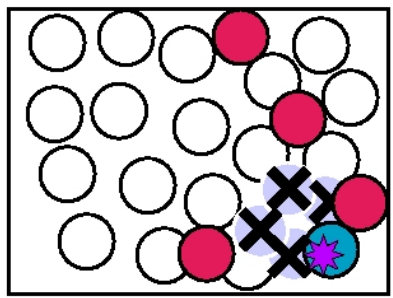

**Step Five:**
Find best matching clone and assign clone's class to antigen. Delete other superfluous clones and for each deletion, replace with new randomly generated antibody. Repeat steps two to five until a stopping condition is met.

Fig. 3 An illustration of clonal selection.

combined with features of K-means where the position of the centroids is adjusted (analogous to the creation of memory antibodies). However, again, the affinity met- rics and the hypermutation components make CSA somewhat distinct from these methods.

The primary uses of CSA are for pattern recognition and optimisation, exemplified by the successful application of an optimised variant of CLONALG termed Opt-AI [18] to the prediction of protein secondary structure. It is the hypermuta- tion component of CSA, where a dynamic local search is performed, that implies its suitability for optimisation, and this is exemplified with Opt-IA. Another example of a CSA is AIRS, a successful multi-class classifier that contains a clonal selection component. This system also employs memory cells, created when a stimulated B- cell has a sustained affinity. Immune memory models frequently accompany clonal selection approaches, but the underlying immunology is rather unclear, even regard- ing the existence of such cells. This has made the development of specific models very difficult but possible, as demonstrated by Wilson et al. [63] with the motif- tracking algorithm.

The process of repeated filtering of candidate solutions in the form of antibody populations results in a type of optimisation when taken within an AIS context, al- though it is debatable whether the solutions provided by the human immune system itself are optimal. As argued by Timmis and Hart [54], CSA has produced solutions that have a tendency to be the most robust, though not necessarily the most optimal. This makes them particularly suited for more complex optimisation problems such as multi-objective optimisation. Their robustness, coupled with the fact that they are one of the most well-understood of the AIS algorithms, makes them a popular choice amongst similar techniques.

## 3.4 Idiotypic Network Approaches

In this section, the basic principles of the idiotypic network theory proposed by Jerne [35] are explained, and a particular example of a hybrid system that combines an immune network with a clonal selection-based model is presented and discussed. A more detailed example of an artificial idiotypic network is provided in the case study in section 4.

In order to understand the principles of the idiotypic network theory, it is necessary to introduce the concepts of epitopes, paratopes and idiotopes. The clonal selection theory states that division occurs for B-cells with receptors that have a high degree of match to a stimulating antigen's binding region or epitope pattern, and that these cells then mature into plasma cells that secrete the matching receptors or antibodies into the bloodstream. Once in the bloodstream the antibody combin- ing sites or paratopes bind to the antigen epitopes, causing other cells to assist with elimination. Antibody paratopes and antigen epitopes are hence complementary and are analogous to keys and locks. Paratopes can be viewed as master keys that may open a set of locks, and some locks can be opened by more than one key.

However, Jerne's network theory suggests that antibodies also possess a set of epitopes and so are capable of being recognized by other antibodies. Epitopes unique to an antibody type are termed idiotopes, and the group of antibodies shar- ing the same idiotope belongs to the same idiotype. When an antibody's idiotope is recognized by the paratopes of other antibodies, it is suppressed and its concentra- tion is reduced. However, when an antibody's paratope recognizes the idiotopes of other antibodies, or the epitopes of antigens, it is stimulated and its concentration increases.

The idiotypic network theory hence views the immune system as a complex net- work of paratopes that recognize idiotopes and idiotopes that are recognized by paratopes. This implies that B-cells are not isolated, but are communicating with each other via collective dynamic network interactions. The network continually adapts itself, maintaining a steady state that reflects the global results of interacting with the environment. This is in contrast to the clonal selection theory, which sup- ports the view that promotion of a B-cell to a memory cell is the result of antibody- antigen interactions only. Jerne states that each individual develops a unique, self- regulating immune network, and when it is established, it must possess stable fea- tures. He hence proposes that immunological memory may be more dependent upon network changes than upon the endurance of populations of memory cells.

His theory asserts that antibodies continue communicating even in the absence of antigens, which produces continual change of concentration levels. A more re- cent model by Farmer et al. [22] adds additional dynamics that account for the domination of a single antibody in the presence of antigen, since the cell with the paratope that best fits the antigen epitope contributes more to the collective response. It presents itself to the system as the antigenic antibody, which disturbs the network, inducing further interantibody suppression and stimulation.

Although the theory has been largely ignored by the wider immunology commu- nity, it has gained much popularity within AIS due to its ability to produce flexible selection-mechanisms. Furthermore, the behaviour of an idiotypic network can be considered intelligent, as it is both adaptive at a local level, and shows emergent properties at a global level. The system is also autonomous and completely decen- tralized, making it ideal for applications such as mobile-robot behaviour arbitra- tion [41][40][59], identifying good matches for recommendation software [13], and negotiating options for configuring communication software [53].

An early example of AIS research inspired by Jerne's theory is the system named 'Jisys', developed by Hunt and Cooke, and later Timmis [33]. The system was based on the idiotypic network model formalised by Farmer et al. [22] and later Bersini and Varela et al. [9], and utilised the concepts of stimulation and suppression effects within a network of antibodies. The system can be considered as something of a hybrid, since it also incorporates the concepts of clonal expansion and somatic hy- permutation within the antibody populations. The system led to the development of a number of other network-based systems, including ANNIE/RAIN [55], which is a resource-based unsupervised clustering algorithm and AINET [19]. Components from ANNIE/RAIN are incorporated into AIRS in addition to elements of clonal selection.

## 3.5 Danger Theory Approaches

All of the algorithms described above (clonal and negative selection, and the immune networks) diverged from the underlying immunology at an early stage in their development. This phenomenon often occurs in AIS because, as novel variants are created, any remaining immune inspiration is abstracted away in order to produce systems that are easy to characterise computationally. Consequently, the resulting systems may fail to model certain computationally desirable features of the immune system. In addition, since the algorithms are developed from a computational perspective, it can be difficult to distinguish AIS approaches from more established machine learning techniques. This is exemplified by the similarities of CSAs with K-nearest neighbour approaches and evolutionary search techniques. Although the first-generation algorithms continue to be applied to numerous pattern recognition, detection and classification problems, little progress has been made with the algo- rithms themselves for a number of years. This, coupled with the somewhat mediocre performances achieved by such algorithms on benchmark tests, has recently led AIS researchers to re-think the fundamentals of AIS design [54]. Instead of using highly-simplified models of isolated immune components, systems could be designed to incorporate more complex, current and sophisticated models. The idea gave rise to a hypothesis; would the incorporation of finer-grained models improve the performance of AIS algorithms, and make them more applicable?

The Danger Project (Aickelin et al. [2]), a four year interdisciplinary collaboration between an AIS development team and practical immunologists aimed to answer this fundamental question. Their research was chiefly motivated by the scal- ing and false positive problems associated with negative selection and was based on a proof of concept paper by Aickelin and Cayzer [3]. Here, the immune system was re-examined in an attempt to overcome the difficulties, and it was postulated that negative selection-based intrusion detection systems may be missing a key con- stituent; danger signals.

As described in section 2, the human immune system cannot rely on self-nonself discrimination alone, so it seems unreasonable to design AIS systems that depend only upon this principle. The aim of the Danger Project was to incorporate the danger theory into AIS, with a view to producing robust intrusion detection sys- tems, capable of fast real-time analysis and low rates of false alarms. At the start of the project in 2004, the working methodology of the research team was unique in AIS; the practical immunologists gave the computer scientists insight into the actual mechanisms of detection employed by the immune system, and the AIS researchers were able to build abstract computational models of the cells involved in the detec- tion of danger signals, which formed the basis of novel algorithms and frameworks. Moreover, the practical immunologists were able to assist in refining the models by performing experiments to fill in any gaps in knowledge that were identified.

Two separate areas of research arose out of the danger project in addition to the published immunological results. The first was the development of the libtissue system, an agent-based framework that facilitated a style of agent-based simulation to house the novel algorithms [57]. A novel algorithm (termed 'tlr') was developed to test the framework, and showed some success when applied to the detection of anomalous system calls [56]. The algorithm is one of the few instances of AIS where more than one cell type is employed, in this case, DCs and T-cells. The second research area was the creation of the DCA [26], a second-generation example, and the newest addition to the mainstream set of AIS algorithms. The DCA is based on a model of the function of dermal dendritic cells and their ability to discriminate between healthy and infected tissue. In nature DCs correlate molecular signals found within tissue and use this information to assess the context of the monitored area. In addition to signal processing, DCs collect debris, which is processed to form antigen. After a period of time, DCs mature and migrate from the tissue to a lymph node, where they present their context information and their antigen to a population of T-cells, instructing the T-cell with the appropriate response.

In the DCA, the DC mechanisms are abstracted and used to form the model. To date the DCA has been applied to port scan detection, insider attack detection, botnet zombie machine detection, standard machine learning intrusion datasets, robotic se- curity, schedule overrun detection in embedded systems, sensor networks, and other real-time, dynamic problems. In numerous cases the algorithm is performing well, producing low rates of false positives, and a deterministic variant that has enhanced computational performance is currently under investigation [28]. The algorithm is described in detail in section 5.

## 3.6 Conceptual Framework Approaches

In parallel with the Danger Project, Stepney et al. [51] also identify the lack of rigour in the metaphors used to inspire AIS. To overcome this problem, they propose a framework (the 'Conceptual Framework') for the successful development of AIS. The methodology employs an iterative approach for the creation and testing of novel immune-inspired algorithms, and four stages are identified as key:

- Observation: the biological system is probed using practical experimentation.
- Models: computational models are constructed to examine the biological sys- tem further, and abstract models are created from the computational models for translation into algorithms.
- Algorithms: computational systems are developed, implemented, and studied the- oretically using the abstract models as a blueprint.
- Applications: the developed algorithms are applied to specific problems, with feedback to the algorithm for refinement.

The design of the framework stipulates that the flow of information between com- ponents is bi-directional, and involves an iterative process, updating the models and algorithms as information is incorporated. A framework for constructing algorithms is certainly necessary in principle, since it clearly defines the role each discipline must play, i.e. observation by immunologists, modelling by mathematicians, algo- rithm development by computer scientists, and application testing by engineers.

Models of receptor degeneracy by Andrews and Timmis [4] are in development using the Conceptual Framework approach, with one modification, i.e. no direct col- laboration with practical immunologists is formed. Instead, sophisticated immuno- logical literature is used as inspiration to construct a novel computational model. Here, the constructed model is of T-cell activation within a lymph node, and a com- putational model of the interactions between T-cells and antigen presenting cells (e.g. DCs) is implemented using principles of cellular automata. In this work, it is identified that one key feature of activation is the degeneracy of receptors across the T-cell population. Degeneracy is defined as "elements which are structurally dif- ferent but produce the same function...". For example, one particular T-cell receptor can respond to more than one binding agent with similar effects. Degeneracy is a de- sirable property that is inherent in numerous biological systems, and is of particular interest to AIS as it may enable reduction of the number of detectors required. This would impact on the dynamics of the first-generation approaches, negative selection included.

The initial model is extended to incorporate tuneable activation thresholds for the responses of T-cells [5]. Dynamic thresholds are employed based on an exist- ing immunological model, where the signal strength needed to activate the T-cell is derived from the frequency and magnitude of the stimulation of the cell over time. Similar research into formalising threshold methods using a type of process algebra known as stochastic $\pi$ calculus has also been carried out, and allows for the formu- lation of models within a defined modelling language [45], utilising the Conceptual Framework for its development.

Both approaches have yielded immunologically and computationally interesting results. However, neither technique has matured to the stage of a workable algorithm and, thus, their applicability to the wider AIS context is still undetermined. It is hoped that the integration of these mechanisms will impact on the function of AIS at some point, stimulating others to follow the Conceptual Framework. However, at the present time, no realistic claims about its effectiveness can be made, as it is too new to have mature work associated with it.

## 3.7 Summary

AIS is a diverse field of study within bio-inspired computation, with the algorithms developed as distinct as the various parts of the immune system itself. This results in not one single AIS algorithm, but a collection of algorithms fit to solve a wide range of problems. Two generations of AIS are currently in use and development. The first-generation approaches draw inspiration from theoretical immunology models in combination with 'text-book style' mechanisms. Two major techniques from the first-generation, clonal and negative selection, share properties with other machine learning methods, such as K-nearest neighbour. Recently, second-generation algo- rithms have emerged, based on an interdiciplinary methodology. Although these approaches are still in the early stages of development, preliminary results, and the increasing popularity of algorithms like the DCA, suggest that second-generation algorithms may prove extremely useful.

To reinforce the concepts presented in this section, two examples of immune-inspired algorithms are examined in more detail. These are the idiotypic network and the DCA, representing first and second-generation algorithms respectively.

## 4 Case Study 1: The Idiotypic Network Approach

Systems inspired by the idiotypic network theory include the interaction between antibodies, in addition to interactions between antibodies and antigens. Such sys- tems are computationally useful, despite the fact that no immunological evidence exists to support the underlying principles. Idiotypic network-based systems are largely inspired by the Farmer et al. computational model [22] of Jerne's idiotypic network theory [35], where binary strings of a given length l represent epitopes and paratopes. The model simplifies the biology so that each antigen and each antibody have only one epitope. Each antibody thus has a pair of binary strings [p, e], and each antigen has a single string [e]. The degree of fit between epitope and paratope strings is analogous to the affinity between real epitopes and paratopes, and uses the exclusive OR operator to test the bits of the strings (where 0 and 1 yields a positive score).

Exact matching between p and e is not required and, as strings can match in any alignment, one needs only to define a threshold value s below which there is no reaction. For example if s was set at 6 and there were 5 matches (0 and 1 pairs) for a given alignment, the score for that alignment would be 0. If there were 6 matches, the score would be 1 and if there were 7 the score would be 2. The strength of reaction G for a given alignment is thus:

$$G = 1 + \mu, \qquad (1)$$

where $\mu$ is the number of matching bits in excess of the threshold. The strength of reaction for all possible alignments $m_{ij}$ between two antibodies i and j is given by:

$$m_{ij} = \sum_{k=1}^{q} G_k, \qquad (2)$$

where q is the number of possible alignments. In the Farmer model, differential equation (3) describes continuous antibody concentration changes occurring as a result of antigen stimulation, interantibody stimulation and suppression, and the nat- ural death rate. Here, N is the number of antibodies and n is the number of antigens. The match specificities are given by m, with the first index referring to the epitope and the second to the paratope:

$$\frac{dx_i}{dt} = c\left[\sum_{j=1}^{n} m_{ij} x_i y_j - k_1 \sum_{j=1}^{N} m_{ij} x_i x_j + \sum_{j=1}^{N} m_{ji} x_i x_j\right] - k_2 x_i, \qquad (3)$$

The first sum in the square bracket expresses stimulation in response to all anti-gens. The $x_i y_j$ terms model that the probability of a potential collision (i.e. match) between an antibody and an antigen (and hence the probability of stimulation) is dependent on their relative concentrations. The second and third sums represent suppression and stimulation respectively, in response to all other antibodies. Parameter $k_1$ allows possible inequalities between interantibody stimulation and suppression, and the $k_2$ term outside the brackets is a damping factor, which denotes the tendency of antibodies to die, at a constant rate, in the absence of interactions. Parameter c is a "rate" constant that simulates both the number of collisions per unit time and the rate of antibody production when a collision occurs. After each iteration, anti-body concentrations are usually reduced using some sort of squashing function, for example the one shown in equation (4):

$$x_i(t+1) = \frac{1}{1 + \exp(0.5 - x(t+1)_i)} \quad (4)$$

The final antibody selected to tackle the presented antigen is the antibody with highest concentration or the best antibody according to a metric that encompasses concentration and affinity.

Within mobile robotics, many researchers (e.g. Watanabe et al. [59]) encode anti-gens as environmental signals, and antibodies as robot behaviours, with the epitopes, paratopes and idiotopes encoded as binary strings. The antibodies have an action and a precondition (the paratope part), which are taken from fixed sets of actions and preconditions. They also have an idiotope part - a disallowed condition, which defines antibody connection. The final antibody structures are determined using a genetic algorithm that evolves suitable combinations of idiotope parameters, actions and preconditions. Both Farmer et al. [22] and Vargas et al. [58] have likened this technique to learning classifier systems (LCS) where antibodies can be thought of as classifiers with a condition and action part (the paratope) and a connection part (the idiotope). In LCS the action part must be matched to a condition (antigen epitope) and the connections show how the classifier is linked to the others. The presence of environmental conditions causes variations in classifier concentration levels in the same way that antigens disturb the antibody dynamics.

In contrast to the binary coding techniques, the work of Whitbrook et al. [61], which is concerned with mobile robot navigation, uses a fixed idiotope matrix of real numbers I representing the degree of belief that an antibody-antigen combination is poor. A variable paratope-matrix of real numbers P, derived from antibody-antigen reinforcement learning (RL) scores, is used to model antibody-antigen affinities. Later work by these authors [62] deals with a set of N antibodies, where each is associated with a particular antigen, i.e. $N = n$. The technique evolves the action part of each antibody in the set and there are z sets. There are thus z matching antibodies for each antigen. The variable paratope is determined by RL as before, but the idiotope is derived directly from the paratope, and is also variable.

P and I are used together to assess similarity between antibodies and hence determine interantibody stimulation and suppression levels. The antibody with the highest affinity to the presented antigen v is selected as the antigenic antibody, i.e.

the antibody with the highest paratope value $P_{iv}$, $i = 1, ..., z$. The system works by suppressing antibodies dissimilar to the antigenic antibody, and stimulating similar ones. This is done by comparing the idiotope of the antigenic antibody with the paratopes of the other antibodies to determine how much each is stimulated, and by comparing the antigenic paratope with the idiotopes of the others to calculate how much each should be suppressed. If the antigenic antibody is the rth antibody and n is the number of antigens, equations (5) and (6) govern the increase $\varepsilon$ and decrease $\delta$ in affinity to v for each of the z matching antibodies, when stimulation and suppression occur respectively:

$$\varepsilon_{iv} = \sum_{j=1}^{n}(1-P_{ij})I_{rj}x_{ij}x_{rj} \qquad i = 1, ..., z, \qquad (5)$$

$$\delta_{iv} = k_1 \sum_{j=1}^{n} P_{rj}I_{ij}x_{ij}x_{rj} \qquad i = 1, ..., z \qquad (6)$$

The new affinity $(P_{iv})_2$ is hence given by:

$$(P_{iv})_2 = (P_{iv})_1 + \varepsilon_{iv} - \delta_{iv}. \qquad (7)$$

All concentrations are re-evaluated using a variation of Farmer's equation, and the antibody selected is the one with the highest activation, which is a product of concentration and affinity. The chosen antibody may be the antigenic antibody, or it may be some other that matches the presented antigen, in which case an idiotypic difference is said to occur. The research has so far shown that both real and virtual robots can navigate through mazes and other obstacle courses much more success- fully when they employ the idiotypic selection mechanism, as opposed to relying on RL only. The authors have also attempted to examine the relationships between the parameters $k_1$, $k_2$, and c, and the rates of idiotypic difference in order to gain insight into the mechansims that underlie the algorithm's superior performance.

## 5 Case Study 2: The Dendritic Cell Algorithm (DCA)

The DCA, a second-generation algorithm based on an abstract model of natural DCs, is one of the most recent additions to the AIS family. It is essentially a meta- heuristic that uses input signals (heuristic approximations of what is normal and anomalous) to perform context-sensitive anomaly detection through both correla- tion and classification. The primary use to date has been the detection of intrusions in the fields of network [27] and robotic security [44].

Natural DCs are part of the innate immune system, and are responsible for ini- tial pathogen detection, acting as an interface between the innate and adaptive sys- tems. They exist in three states of differentiation; immature, mature (exposed to the molecules from necrosing cells), and semi-mature (exposed to the molecules from apoptosing cells), and it is their terminal state of differentiation that is used by the

Artificial Immune Systems                                                                 23adaptive immune system to decide whether or not to respond to a potentially harmful entity.

The DCA abstracts a multi-resolution model of a natural DC. To this end, four data types are required:

- Antigen: a enumerated type object with a value that is used as an identifier for the suspect data to be classified. For ideal functioning, a number of antigens of the same value should be used to form an antigen type.
- PAMP signal: a real-valued attribute, where an increase in value is a definite indicator of anomaly.
- Danger signal: a real-valued attribute, where an increase in value is a probable indicator of damage, but there is less certainty than with a PAMP signal.
- Safe signal: a real-valued attribute, where an increase in value is a probable indicator of normality within a system. High values of the safe signal can cancel out the effects of both the PAMP and danger signals, possibly reducing the false positive rate of the DCA.

Unlike the negative selection approaches, the DCA does not have an adaptive component and thus requires no formal training phase. Signal processing, the corre- lation between antigen and signals, is performed at the individual cell level, but the classification of antigen types occurs at the population level. In other words, each cell's input signals are transformed into cumulative output signals, with two output values per cell; the costimulatory signal (CSM), and the context value k, which is used to determine the terminal state of the cell. (A negative k represents a semi- mature cell and a positive k indicates a mature cell.) Each DC executes three steps per sampling iteration:

1. Sample antigen: the DC collects antigen from an external source (in the form of an antigen array) and places the antigen in its own data structure for storage.
2. Update input signals: the DC collects values of all input signals from a defined input signal array.
3. Calculate interim output signals: at each iteration, each DC generates three temporary output signal values from the received input signals, and these output values are added to obtain the cell's CSM and k values.

The DCA works on a static population of cells in which every cell removed is immediately replaced. Diversity is maintained across the population by random allocation of lifespan values within a specified range, i.e. when a cell is created it is given a limited time window for data sampling. This is also thought to add robustness to the algorithm. However, the lifespan reduces whenever the CSM value increases, since the CSM value is automatically derived from it. Once the lifespan is over, the cell ceases its sampling iterations and presents all the collected antigens, so that its k value can be determined. The cell is then reset and returned to the sampling iterations with a new lifespan value.

The antigens presented by each cell are processed across all presentations by all members of the DC population. The anomaly score, $K_\alpha$, of an antigen type is cal- culated using the k values presented in conjunction with each antigen type. Initial implementations of the DCA [30] are based on numerous probabilistic components, including random sorting of the DC population, probabilities of antigen collection, decay rates of signals, and numerous other parameters. While the same research shows that the algorithm can be successfully applied to real-world intrusion data, the system is very difficult to analyse given the sheer numbers of probabilities and parameters. Recently, a deterministic DCA [28] has been developed in order to ob- tain a greater understanding of the algorithm's function. The pseudocode for the deterministic version of DCA is given in Algorithm 1.

```
input : Antigen and Signals
output: Antigen Types and cumulative k values
set number of cells;
initialise DCs();
while data do
    switch input do
        case antigen
            antigenCounter++;
            cell index = antigen counter modulus number of cells ;
            DC of cell index assigned antigen;
            update DC's antigen profile;
        end
        case signals
            calculate csm and k;
            for all DCs do
                DC.lifespan -= csm;
                DC.k += k;
                if DC.lifespan <= 0 then
                    log DC.k, number of antigen and cell iterations ;
                    reset DC();
                end
            end
        end
    end
end
for each antigen Type do
    calculate anomaly metrics;
end
```

Algorithm 1: Pseudocode of the deterministic DCA.

While the DCA removes the need to define self, it is still necessary to select a suitable antigen representation and to perform pre-classification of input signals. This represents a fundamental difference between this approach and negative selec- tion for example, as the DCA relies on heuristic-based signals that are not absolute representations of normal or anomalous. Although it bears an intitial resemblance to neural networks, the variable lifespan, the population dynamics and the combined functionality of filtering, correlation and classification set it apart from these other approaches. Full details of the DCA are presented in Greensmith [26] and Green- smith et al. [29].

## 6 Conclusions

This chapter has examined AIS from the underlying immunology and the controversy surrounding competing theories, to the application and implementation of im- mune algorithms as exemplified by the DCA. As mentioned, AIS are distinct from other fields within bio-inspired computing as not one archetypal system is used, but different methods are employed for different purposes. There are two reasons for this; first, the immune system is itself multifunctional, performing many different natural computational tasks, and second, AIS researchers model the immune sys- tem in different ways to suit their goals. For example, clonal selection with its hy- permutation function performs a type of local search and can be modified to perform optimisation. Alternatively, the DCA is based on natural DCs, which are responsible for initial pathogen detection in tissue and can hence be used for anomaly detection.

A shift is also evident in the methods used to create and develop AIS algorithms, from somewhat simplistic models based on out-dated immunology to sophisticated interdisciplinary approaches based on rigourous abstract modelling, see the Danger Project [2] and Conceptual Framework [51]. While the Conceptual Framework approaches are not yet mature enough to yield tangible results, the DCA (developed using this method) is performing well across a range of problems [29] in comparison with other nature-inspired techniques.

### 6.1 Future Trends in AIS

The percentage of research directed towards specific areas of AIS can be estimated by looking at the AIS work reported in the 2008 ICARIS annual conference [8]. Exactly half of the papers in these proceedings are related to the application of first-generation algorithms and their variants. The second largest category is that of the second-generation approaches, followed by theoretical studies of AIS and computational immunology. This differs vastly from the state of the field a mere five years ago, where the two largest groups of papers were applications of clonal and negative selection respectively, followed by idiotypic networks. This change in focus of the field suggests that, as the characterisation of the second-generation approaches improves, they will increase in popularity and may eventually dominate the field. What the future holds for AIS, like any discipline, is uncertain given that AIS algorithms are still evolving. As our knowledge of immunology increases, at some point in the future we may have the grounding and computational resources to build full, biologically accurate computational immune systems, based on both the innate and adaptive systems and their numerous cell types.

## Acknowledgements

This work is supported by the EPSRC (EP/D071976/1). The authors would like to thank Jon Timmis for his feedback and assistance.